\documentclass{article}

\usepackage{microtype}
\usepackage{graphicx}
\usepackage{booktabs} 
\usepackage{amsmath}
\usepackage{amssymb}
\usepackage{amsfonts}
\usepackage{multirow}
\usepackage{xurl}
\usepackage{breakurl}
\usepackage[breaklinks]{hyperref}

\usepackage[accepted]{icml2021}

\icmltitlerunning{Source-Criticism Debiasing}

\begin{document}

\twocolumn[
\icmltitle{A Source-Criticism Debiasing Method \\
for GloVe Embeddings}




\begin{icmlauthorlist}
\icmlauthor{Hope McGovern}{cam}
\end{icmlauthorlist}

\icmlaffiliation{cam}{Cambridge Computer Laboratory, University of Cambridge, Cambridge, United Kingdom}

\icmlcorrespondingauthor{Hope McGovern}{hem52@cam.ac.uk}

\icmlkeywords{Machine Learning, Bias, Influence Functions, Word Embeddings, GloVe, ICML}

\vskip 0.3in
]



\printAffiliationsAndNotice{}  

\begin{abstract}
It is well-documented that word embeddings trained on large public corpora consistently exhibit known human social biases. Although many methods for debiasing exist, almost all fixate on completely eliminating biased information from the embeddings and often diminish training set size in the process. In this paper, we present a simple yet effective method for debiasing GloVe word embeddings \cite{pennington2014glove} which works by incorporating explicit information about training set bias rather than removing biased data outright. Our method runs quickly and efficiently with the help of a fast bias gradient approximation method from \citet{Brunet2019}. As our approach is akin to the notion of ‘source criticism' in the humanities, we term our method Source-Critical GloVe (SC-GloVe).  We show that SC-GloVe reduces the effect size on Word Embedding Association Test (WEAT) sets without sacrificing training data or TOP-1 performance. 
\end{abstract}

\section{Introduction}
\label{submission}
Although many debiasing methods have been proposed to combat the problem of undesirable word associations in embeddings, each one suffers particular drawbacks and might ultimately succeed only in hiding the unsavory associations in latent space \cite{Gonen2019}. Debiasing remains a challenge because it is not well understood how individual training data or subsets thereof influence models in downstream tasks. However, recent work in applying influence functions to word embeddings has made it computationally tractable to identify how much each individual training example influences the overall bias of the model at inference time \cite{Guo2020, Chen2020a}. We make use of this development to re-embed GloVe word vectors by artificially scaling the co-occurrence matrix such that the model learns stronger word relationships from documents which are not biased with respect to some predefined bias metric.\par  

Our approach is based on the notion that an ideal debiased model is not one which has no conception of bias, but rather one that understands its own bias and therefore can self-correct. This is essentially the inclusion of an explicit bias representation. In the humanities, source criticism is a method of evaluating the contextual lens of an informational source in order to determine its reliability, and we apply that concept here. Computationally, we use a method of approximating differential bias from \citet{Brunet2019} to generate a weighting factor for each document which corresponds to how much it affects downstream bias at test time. This is accomplished with a single pass through the corpus. We then use these weighting factors to update the word vectors relevant to our bias metric to what they would have been had the biased document had been counted as `less reliable' during training. Our method is simple, elegant, and represents a novel approach of debiasing via explicit bias inclusion.

\section{Background}
\subsection{Bias in Word Embeddings}
Word embeddings are used widely in natural language processing (NLP) for a broad range of downstream tasks. However, as \citet{Caliskan2017} and \citet{Garg2018} have shown, these embeddings confirm many human social biases, some of which lead to problematic behaviors of machine learning models which rely on pretrained embeddings \cite{Kiritchenko}. Among popular embedding models are Word2Vec \cite{Mikolov}, GloVe \cite{pennington2014glove}, and FastText \cite{Joulin2017}. All three use unsupervised learning techniques on billions of words sourced from large internet corpora. \citet{Bolukbasi2016} demonstrated that a simple analogy task could be used to reveal unwanted latent semantic associations in the embeddings; for example, the response from pretrained GloVe embeddings to the question ``man is to computer programmer as woman is to X'' rendered ``homemaker'' as the most likely answer. \par
\citet{Caliskan2017} introduced a standardized method of measuring biases present in word embeddings from a template of the Implicit Association Test (IAT), which is used widely in the field of sociology to measure latent human prejudices; this includes everything from morally neutral biases, such as `flowers are more pleasant than insects' to more problematic ones, such as `European names are more pleasant than African names'. From this, they derive the Word Embedding Association Test (WEAT), which gives the probability that the observed similarity scores could have arisen with no semantic association between the target concepts and the attribute. \par
Some previously proposed debiasing methods happen post-training, such as \citet{Bolukbasi2016}, focus on zeroing the `gender direction', i.e. the projection of each word on a predefined gender direction written as $\overrightarrow{\textbf{w}} \cdot (\overrightarrow{\textbf{he}} - \overrightarrow{\textbf{she}}) = 0$. Other methods attempt to mitigate bias during training, such as \citet{Zhao2018}'s method of encouraging gendered information to reside in the tail end of the word vector during training and then removing the biased subset of the word vector before use in an NLP system. 

\subsection{Influence Functions}
With the use of influence functions, a methodology borrowed from robust statistics, it is possible to determine which inputs to a model exert the most influence over model inference at test time \cite{Koh2017}. Fundamentally, influence functions are an approximation of the result that would be achieved by removing one example at a time from the dataset and training a model to see its net effect on inference. Recent work in this area has introduced computationally efficient tools for performing fast influence function calculations for large models \cite{Guo2020} and even through a multi-stage training (pre-training and fine-tuning) process \cite{Chen2020a}. These tools provide insight into model behavior by directly tracking how single training examples affect downstream inference and therefore hold potential for the growing trend of explainable machine learning \cite{bhatt2020explainable}.\par
\citet{Brunet2019} apply an influence function based approximation method to word embeddings as a way of explaining learned bias. With their method, it is possible to determine the subsets or individual documents that are most responsible for disparate gendered representations in learned word vectors. They demonstrate this method by creating perturbation sets which remove a number of biasing documents and compare the predicted differential bias sets with ground truth removal and re-embedding. They show remarkable agreement between the two, giving strong evidence that their approximation function is a trustworthy proxy for differential bias. 

\section{Methodology}
\subsection{Choice of Bias Metric}
For our experiments, we consider the \textit{effect size} of two different WEAT bias word sets as introduced by \citet{Caliskan2017}.The WEAT test itself measures the similarity of words $a$ and $b$ in word embedding $w$ as measured by the cosine similarity of their vectors, $\cos{(w_a, w_b)}$. In WEAT1, the target word sets relate to \textit{science} and \textit{arts} terms, with the expectation that scientific terms will cluster more with `male' attributes sets and art terms will cluster more with `female' attribute sets. In WEAT2, the target word sets relate to \textit{weapons} and \textit{instruments}, with the expectation that weaponry terms will cluster more with `unpleasant' attribute sets and musical terms will cluster more with `pleasant' attribute sets. A representative list of the word sets is seen in Table \ref{tab:wordlists}. Of the bias sets used, WEAT1 is broadly considered to reflect a problematic bias whereas WEAT2 reflects one that is more benign. 

\begin{table*}[h]
\centering
\begin{tabular}{|p{1cm}||p{7cm}||p{7cm}|}
    \hline 
    \textbf{WEAT} & \textbf{Target Sets} & \textbf{Attribute Sets} \\ 
    \hline
    \multirow{2}{*}{I}& \small \textbf{science:} science, technology, physics, chemistry, einstein, nasa, experiment, astronomy
    & \small \textbf{male:} male, man, boy, brother, he, him, his, son \\
    \cline{2-3}
     & \small \textbf{arts:} poetry, art, shakespeare, dance, literature, novel, symphony, drama& \small \textbf{female:} female, woman, girl, sister, she, her, hers, daughter \\
    \cline{2-3}
    \hline
    \multirow{2}{*}{II}&\small \textbf{instruments:} bagpipe, cello, guitar, lute, trombone, banjo, clarinet, harmonica, mandolin, trumpet, bassoon, drum, harp, oboe, tuba, bell, fiddle, harpsichord, piano, viola, bongo, flute, horn, saxophone, violin\ & \small\textbf{pleasant:} caress, freedom, health, love, peace, cheer, friend, heaven, loyal, pleasure, diamond, gentle, honest, lucky, rainbow, diploma, gift, honor, miracle, sunrise, family, happy, laughter, paradise, vacation \\
    \cline{2-3}
    & \small \textbf{weapons:} arrow, club, gun, missile, spear, axe, dagger, harpoon, pistol, sword, blade, dynamite, hatchet, rifle, tank, bomb, firearm, knife, shotgun, teargas, cannon, grenade, mace, slingshot, whip &\small \textbf{unpleasant:} abuse, crash, filth, murder, sickness, accident, death, grief, poison, stink, assault, disaster, hatred, pollute, tragedy, divorce, jail, poverty, ugly, cancer, kill, rotten, vomit, agony, prison\\
    \cline{2-3}
    \hline
\end{tabular}
\caption{Full WEAT word lists}
\label{tab:wordlists}
\end{table*}

\subsection{GloVe}
We focus on only one kind of embedding model, GloVe, in this paper, although our method could theoretically be applied to others. GloVe (Global Vectors for word representations) is essentially a log-bilinear model with a weighted least-squares objective \cite{pennington2014glove}. The underlying intuition for the model is that words which occur in the same context more frequently within a given corpus are more likely to be semantically linked.
The training objective of GloVe is to learn word vectors such that their dot product equals the logarithm of the words' probability of co-occurrence.\par
Formally, this happens in two steps\footnote{In the following sections, we use the mathematical notion provided in \citet{Brunet2019}}. In the first, a sparse co-occurrence matrix $X \in \mathbb{R}^{V \times V}$ is extracted from the corpus. Each entry in the matrix, $X_{ij}$ represents a weighted count of the number of times that word $j$ occurs in the context window of word $i$. For the second step, the optimal embedding parameters $w^*, u^*, b^*,$ and $c^*$ are learned via gradient-based optimization such that they minimize the loss: 
\begin{equation}
    \small J=\sum_{i=1}^{V}\sum_{j=1}^{V} f(X_{ij})(w_i^T u_j + b_i + c_j - \log{X_{ij}})^2
\end{equation}
where $w_i \in \mathbb{R}^{D}$ is the embedding of the $i$th word in the vocabulary. We use an embedding dimension of 75. The set of $u_j \in \mathbb{R}^{D}$ are the context word vectors, and $b_i$ and $c_j$ are the bias terms for $w_i$ and $u_j$, respectively. $f(x)$, the weighting function, is used to attribute more importance to common word occurrences. 

\subsection{Data Description}
Our dataset consists of a corpus constructed from a Simple English Wikipedia dump\footnote{\url{https://dumps.wikimedia.org/ simplewiki/}}, using 75-dimensional word embedding vectors. The TOP-1 analogies test (shipped with the GloVe code base) measured approximately 35\%, which is certainly lower than state-of-the-art but high enough for our purposes of demonstrating proof-of-concept of our method. We note that future work would ideally consider a broader range of corpora for testing. We train an initial GloVe embedding model on this corpus; the relevant corpus statistics and GloVe training hyperparameters are listed in Table \ref{tab:corpus}.

\begin{table}
\centering
\begin{tabular}{lc}
\hline 
\multicolumn{2}{c}{\textbf{Wiki}} \\
\hline
\textbf{Corpus} & \\
\hline
Min. doc. length & 200 \\
Max. doc. length & 10,000  \\
Num. documents & 29,344  \\
Num. tokens & 17,033,637 \\
\hline
\textbf{Vocabulary} & \\
\hline
Token min. count & 20 \\
Vocabulary size & 44,806 \\
\hline
\textbf{GloVe} & \\
\hline
Context window & symmetric \\
window size & 8 \\
$\alpha$ & 0.75 \\
$x_{max}$ & 100 \\
Vector Dimension & 75 \\
Training epochs & 300 \\
\hline
\textbf{Performance} & \\
\hline
TOP-1 Analogy & 35\% \\
\hline
\end{tabular}
\caption{\label{tab:corpus} Experimental setup for Wiki corpus.}
\end{table}
\subsection{Differential Bias}
For our debiasing method, it is crucial that we know how much each document in the corpus contributes to its downstream effect size. The naive way to compute this would be to remove a single document from the corpus and completely retrain the embedding, but this is impractical and computationally intractable even for relatively small NLP corpora. However, \citet{Brunet2019} introduce a method for calculating an approximation of the resulting embedding change by applying a modified version of influence functions\footnote{the code for this is publicly available at \url{https://github.com/mebrunet/understanding-bias}}. They approximate that the optimal word vector learned from the initial training, $w^*_i$, will change with respect to a given corpus perturbation (removing a document) as:
\begin{equation}
   \small \tilde{w_i} \approx w^*_i - \frac{1}{V}H^{-1}_{w_i}[\nabla_{w_i}L(\tilde{X_i}, w) - \nabla_{w_i}L(X_i, w)]
   \label{eq:wi}
\end{equation}

Where $\tilde{w_i}$ is the word vector learned from a perturbed corpus, $V$ is the size of the vocabulary, $X$ is the global co-occurrence matrix, and $\tilde{X}$ is the co-occurrence matrix discounting the co-occurrence matrix of document $i$. $H$ is the Hessian with respect to only word vector $w_i$ of the point-wise loss at $X_i$, and
$\nabla_{w_i}L(X_i, w)$ is the gradient of the point-wise loss function at $X_i$ with respect to only word vector $w_i$.

We use this method to get an approximation of the differential bias of each document in the Wiki corpus. Once that is calculated, we pass to our method the co-occurrence matrix, the trained GloVe model, the WEAT test words, and the newly created list of differential biases, where $\beta^{(k)}$ is the differential bias approximation for document $k$.
The full algorithm may be seen in Algorithm 1. It bears similarity to the differential bias approximation in \citet{Brunet2019} with the crucial changes that we weight the subtractor of the co-occurrence matrix by the pre-calculated differential bias, and that we actually set the approximated vector as the vector stored in the final GloVe model. This is because we are approximating what the word vector would have been had it paid less attention to a biased document, so we update the word vector to reflect that change, rather than temporarily storing it for comparison. \par
The intuition for this algorithm is that WEAT words which appear in heavily biased documents will lend the model unsavory associations between those words. Instead, we maximize the strength of the connection between target and attribute words that appear together in unbiased or, better yet, de-biasing documents (those with a negative differential bias value).\par Given a set of WEAT words in a document, there are three possible actions according to our method: if the document does not affect downstream bias with respect to some bias metric, the weighting factor is zero and the true co-occurrence matrix is used. If the document increases bias at test time, the co-occurrence matrix is slightly artificially decreased for the relevant words, scaled by how heavily biased it is. If the document decreases bias downstream, the co-occurrence matrix is slightly increased for the relevant words, artificially strengthening the co-occurrences of more balances terms. We find this method to be an elegant and intuitive approach to debiasing word embeddings. 

\begin{table}
\centering
\begin{tabular}{lrl}
\hline 
\textbf{Algorithm 1} Source-Criticism Debiasing \\
\hline
\textbf{input} \textit{Co-occ Matrix: X, WEAT words: \{S,T, A, B\}} \\
\quad \quad \textit{Diff Bias Vector: $\beta$},\\
\quad \quad $w^*, u^*, b^*, c^*c = GloVe(X)$\\
\quad \textbf{for} doc \textbf{in} corpus \textbf{do}\\
\quad \quad \textit{\# weight the co-occ matrix} \\
\quad \quad $X' = X-(\beta^{(k)}\cdot X^{(k)})$ \\
\quad \quad \textbf{for} word $i$ \textbf{in} doc $\cap(S\cup T\cup A\cup B)$ \textbf{do}\\
\quad \quad \quad \textit{\# approximate WEAT word vecs} \\
\quad \quad \quad $\tilde{w_i} = $ \textit{\# see Equation }\ref{eq:wi}\\
\quad \quad \quad \textit{\# update WEAT word vecs} \\
\quad \quad \quad $w^*_i = \tilde{w_i}$\\
\quad \quad \textbf{end for} \\
\quad \textbf{end for} \\
\textbf{re-evaluate WEAT with new word vectors}\\
\hline
\end{tabular}

\end{table}
\section{Results}
Table \ref{tab:results} shows the baseline WEAT effect sizes compared to our model, SC-GloVe, where the best result is reported in bold. We show that SC-GloVe does decrease the effect size for both WEAT test sets used. Each of these scores was averaged over 10 trials to counteract the variability that is inherent in the optimization process. 
We re-run the built-in TOP-1 analogy test on the final SC-GloVe model and find that it similarly achieves around 35\%.

\begin{table}[h]
\centering
\begin{tabular}{lcc}
\hline 
\textbf{Model} & \textbf{WEAT1} & \textbf{WEAT2} \\
\hline
Baseline & 0.577  & 1.04 \\
SC-GloVe & \textbf{0.461} & \textbf{0.964} \\
\hline
\end{tabular}
\caption{\label{tab:results} WEAT Effect Sizes.}
\end{table}

\section{Discussion}
Our intention in this paper is to present a compelling proof-of-concept for a novel method rather than to fine-tune it for competitive performance against existing debiasing methods. As such as, we simply note that against our two chosen WEAT test sets, our method successfully decreases downstream bias effect size. This is done without removing training data, but by a simple weighting factor and efficient re-embedding algorithm. Encouragingly, it does not decrease TOP-1 analogy performance, which indicates the the method is targets only the offending vectors.\par
One weakness of this approach is that it is highly specific to the chosen bias metric. The source criticism method we introduce must be performed for each of the WEAT word sets, which hinders a blanket applicability to remove all enumerated biases in a GloVe model at once. Future work will seek to expand the coverage of the method to debias with respect to multiple metrics at once, in addition to incorporating statistical significance testing across a variety of different GloVe pre-training parameters and corpora. We are satisfied to present proof-of-concept on a relatively small corpus size in this paper. \par

\section{Conclusion}
We have presented a simple, effective, and conceptually novel method for incorporating explicit bias representation into a word embedding debiasing method. We demonstrate that our model, SC-GloVe, can reduce downstream bias effect sizes with respect to a defined bias metric. We hope to refine this method further with future work in order to demonstrate its value in debiasing word embeddings for NLP models.

\section*{Acknowledgements}
The author would like to acknowledge the Cambridge Computer Laboratory and 
Dr. Marcus Tomalin for providing feedback on an earlier version of this work, as well as Umang Bhatt, who contributed to the ideas contained herein. 




\bibliography{icml2021}
\bibliographystyle{icml2021}

\end{document}